%% file: root.tex
\documentclass[letterpaper, 10 pt, conference]{ieeeconf}  % Comment this line out if you need a4paper

\IEEEoverridecommandlockouts                              % This command is only needed if 
\usepackage{algorithm}
\usepackage{algpseudocode}
\usepackage{array} %%%
\usepackage{graphicx}%%%
\usepackage{amsmath} %%%
\usepackage{comment}
\usepackage{booktabs}       % professional-quality tables
\usepackage{multirow}
\usepackage{makecell}
\usepackage{adjustbox}
\usepackage{xcolor}
\usepackage{cite}

\usepackage{bbold}
\usepackage{amsfonts}
\usepackage{amssymb}
\usepackage{txfonts}
\usepackage{bbm}

\usepackage{booktabs}
\usepackage{nicematrix}

\newcolumntype{C}{>{\centering\arraybackslash}p{4.5em}}

\title{\Large \bf
PGA: Personalizing Grasping Agents with Single Human-Robot Interaction
}
\author{Junghyun Kim$^{12}$ Gi-Cheon Kang$^{12*}$ Jaein Kim$^{13*}$ Seoyun Yang$^{4}$ Minjoon Jung$^{12}$ Byoung-Tak Zhang$^{123}$%
\thanks{$^*$Authors have equal contributions}
\thanks{$^{1}$AI Institute, Seoul National University}
\thanks{$^{2}$Interdisciplinary Program in AI, Seoul National University}
\thanks{$^{3}$Interdisciplinary Program in Neuroscience, Seoul National University}
\thanks{$^{4}$Division of Engineering Science, University of Toronto}
\thanks{
This research was supported by the Institute of Information \& Communications Technology Planning \& Evaluation (IITP) (2021-0-02068-AIHub/10\%, 2021-0-01343-GSAI/30\%, 2022-0-00951-LBA/20\%, 2022-0-00953-PICA/40\%) grant funded by the Korean government.
}
}

\begin{document}

\maketitle
\thispagestyle{empty}
\pagestyle{empty}

%%%%%%%%%%%%%%%%%%%%%%%%%%%%%%%%%%%%%%%%%%%%%%%%%%%%%%%%%%%%%%%%%%%%%%%%%%%%%%%%
\begin{abstract}

Language-Conditioned Robotic Grasping (LCRG) aims to develop robots that comprehend and grasp objects based on natural language instructions. 
While the ability to understand personal objects like \textit{my wallet} facilitates more natural interaction with human users, current LCRG systems only allow generic language instructions, \textit{e.g.,} \textit{the black-colored wallet next to the laptop}.
To this end, we introduce a task scenario \textsc{GraspMine} alongside a novel dataset aimed at pinpointing and grasping personal objects given personal indicators via learning from a single human-robot interaction, rather than a large labeled dataset.
Our proposed method, Personalized Grasping Agent (PGA), addresses \textsc{GraspMine} by leveraging the unlabeled image data of the user's environment, called \textit{Reminiscence}.
Specifically, PGA acquires personal object information by a user presenting a personal object with its associated indicator, followed by PGA inspecting the object by rotating it.
Based on the acquired information, PGA pseudo-labels objects in the \textit{Reminiscence} by our proposed label propagation algorithm.
Harnessing the information acquired from the interactions and the pseudo-labeled objects in the \textit{Reminiscence}, PGA adapts the object grounding model to grasp personal objects.
This results in significant efficiency while previous LCRG systems rely on resource-intensive human annotations---necessitating hundreds of labeled data to learn \textit{my wallet}.
Moreover, PGA outperforms baseline methods across all metrics and even shows comparable performance compared to the fully-supervised method, which learns from 9k annotated data samples.
We further validate PGA's real-world applicability by employing a physical robot to execute \textsc{GrsapMine}.
Code and data are publicly available at https://github.com/JHKim-snu/PGA.

\end{abstract}

% Rephrased Abstract (by mjjung)
% Language-Conditioned Robotic Grasping (LCRG) aims to grounding and grasping objects based on given human instructions.
% While personalization and human-robot interaction are essential in real-world environments, current datasets cannot specifically test for personalization ability. => 기존애들이 이를 고려안함. 미흡함
% To this end, we introduce a novel LCRG dataset, named \textsc{GraspMine}, to evaluate the model's personalization capabilities. {naive하게 씀. 구체적으로 데이터셋, task설명 들어가면 좋을듯, e.g., My obejct를 집는다}
% Through evaluating previous LCRG methods on our new dataset, we empirically demonstrate their lack of personalized capabilities.
% To tackle these challenges, we propose Personalized Grasping Agent (PGA), a robotic framework that enables recognizing personal items through a single human-robot interaction. 
% {PGA 설명 추가}
% Extensive experiments on offline and online settings demonstrate the effectiveness of PGA and its applicability to real-world environments.

%%%%%%%%%%%%%%%%%%%%%%%%%%%%%%%%%%%%%%%%%%%%%%%%%%%%%%%%%%%%%%%%%%%%%%%%%%%%%%%%

%%%%%%%%%%%%%%%%%%%%%%%%%%%%%%%%%%%%%%%%%%%%%%%%%%%%%%%%%%%%%%%%%%%%%%%%%%%%%%%%
\input{01_intro}

\input{02_related_works}
\input{03_method}

\input{04_dataset}

\input{04_experiment}
\input{05_discussion}
\input{06_conclusion}
%%%%%%%%%%%%%%%%%%%%%%%%%%%%%%%%%%%%%%%%%%%%%%%%%%%%%%%%%%%%%%%%%%%%%%%%%%%%%%%%

%\addtolength{\textheight}{-12cm}   % This command serves to balance the column lengths
                                  % on the last page of the document manually. It shortens
                                  % the textheight of the last page by a suitable amount.
                                  % This command does not take effect until the next page
                                  % so it should come on the page before the last. Make
                                  % sure that you do not shorten the textheight too much.

%%%%%%%%%%%%%%%%%%%%%%%%%%%%%%%%%%%%%%%%%%%%%%%%%%%%%%%%%%%%%%%%%%%%%%%%%%%%%%%%

% \section*{Acknowledgements}
%\setstretch{1}

\noindent \textbf{Acknowledgements.} We extend our gratitude to Jungmin Lee for her invaluable contribution to video editing, and all the reviewers for their insightful comments and feedback.

% \clearpage
%\bibliographystyle{ieeetr}
\bibliographystyle{IEEEtran.bst}
\bibliography{ref}

\end{document}

%% file: 01_intro.tex
\section{INTRODUCTION}
\label{sec:intro}

% Paragraph 1: introduction of the task (WRITE why this task is important)

Empowering robots with the ability to comprehend human natural language presents a formidable yet vital challenge within the realms of AI and Robotics~\cite{robots_that_use_language}. 
This capability, which involves understanding and executing human language instructions, allows for more intuitive human-robot interactions.
The researchers have studied such capability in the context of Language-Conditioned Robotic Grasping (LCRG)~\cite{gvcci, srfnlifrm, invigorate, ingress, lopfribhsr, cpaptbgl, avgrefrm, ivgorefhri, reiofitsf, iprwowusli}, which focuses on robotic systems that ground and grasp objects based on language instructions.

\begin{figure}[t]
\centering
\includegraphics[width=\linewidth]{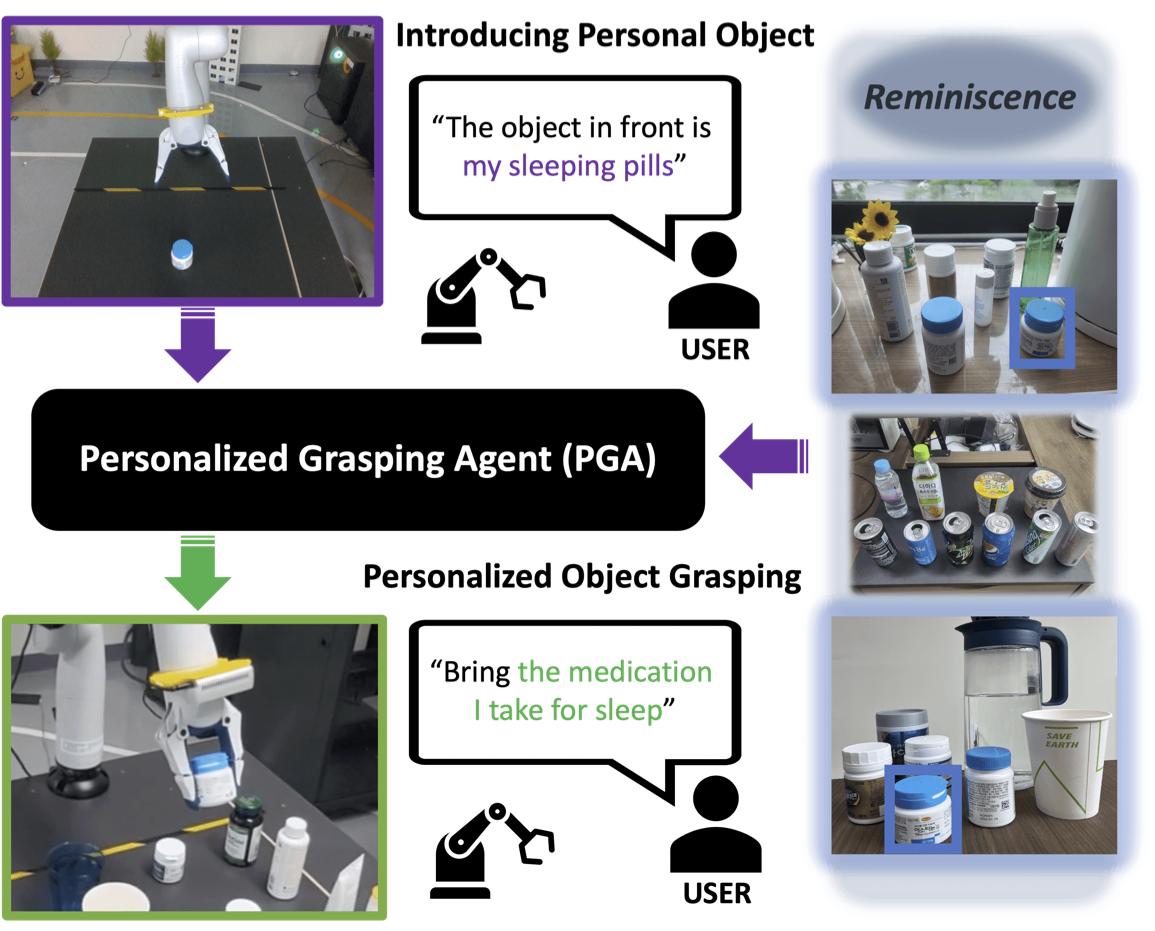}
\caption{
\textbf{Personalizing grasping agents with single human-robot interaction.} 
Upon the user's introduction of a personal object, it retrieves the identical objects from its visual reminiscence.
Leveraging the retrieved objects, the robot subsequently engages in an integrated learning process of the personal object.
The Personalized Grasping Agent (PGA) can finally comprehend and grasp the personal object.
}
\label{fig:first}
\vspace*{-0.5cm}
\end{figure}

Current approaches in LCRG~\cite{srfnlifrm, invigorate, ingress, lopfribhsr, cpaptbgl, avgrefrm, ivgorefhri, reiofitsf, iprwowusli, gvcci} predominantly rely on generic language expressions when describing objects for manipulation, leading to less intuitive human-robot communication.
For instance, a person might instinctively instruct, ``\textit{Get my wallet}''. 
However, with current LCRG systems that operate on generic instructions, one might need to instruct, ``\textit{Get the black-colored wallet next to the laptop}''. 
This forces users to craft their instructions to suit the robot's limited knowledge base, what can be termed as \textit{robot-centric} directives. 
However, many non-expert users find this \textit{robot-centric} directives both unfamiliar and cumbersome. 
To truly facilitate more intuitive \textit{user-centric} interactions, robots should have more shared personal knowledge with their users beyond the generic, understanding personal objects.

To address this problem, we introduce a novel personalized task scenario of LCRG called \textsc{GraspMine} with a benchmark dataset.
\textsc{GraspMine} aims to locate and grasp personal objects given a personal indicator, \textit{e.g., ``my sleeping pills''}, that robotic systems with generic knowledge may not handle properly.
According to the field of Personalization~\cite{fluffy, meta_personalizing}, personalizing with a few examples is a common scenario since collecting a lot of personalized data for each user is infeasible in real-world applications.
Considering this aspect, \textsc{GraspMine} requires robots to learn personal objects via minimal human-robot interaction, \textit{i.e.}, a one-time verbal introduction of a personal object, as depicted in Fig.~\ref{fig:first}.
In \textsc{GraspMine} dataset, we consider learning of 96 personal objects, containing five distinct test splits (Heterogeneous, Homogeneous, Paraphrased, and Cluttered), each sample including an image containing multiple objects, a personal indicator, and the associated object's coordinates.

However, with the prevailing learning strategies of current LCRG systems~\cite{srfnlifrm, invigorate, ingress, lopfribhsr, cpaptbgl, avgrefrm, ivgorefhri, reiofitsf, iprwowusli}, supervised learning that requires vast amounts of data, it is challenging to tackle \textsc{GraspMine} with only a single sample.
To address \textsc{GraspMine}, we propose \textbf{P}ersonalized \textbf{G}rasping \textbf{A}gent (PGA), that learns personal objects by propagating the information acquired from a user through a \textit{Reminiscence}.
Specifically, PGA first constructs the \textit{Reminiscence}---a collection of raw images from the user's environment.
Then PGA acquires personal object information in two successive steps: \textit{human-robot interaction} and \textit{robot-object interaction}. 
In \textit{human-robot interaction}, a user introduces a personal item to the robot with a personal indicator, \textit{e.g.,} ``\textit{my sleeping pills}''.
In \textit{robot-object interaction}, the robot inspects the object from multiple perspectives, obtaining diverse images of personal objects.
Leveraging the acquired information of the personal object, PGA pseudo-labels objects in the \textit{Reminiscence} by propagating personal indicators based on the visual features of the objects, inspired from the label propagation algorithm~\cite{label_propagation}.
We term this \textit{Propagation through Reminiscence}.
Harnessing the data obtained from the interactions and the pseudo-labeled objects in \textit{Reminiscence}, PGA adapts the object grounding model, and with the personalized model, it adeptly grasps the user-specified personal objects.

In the experiment, we provide baselines for \textsc{GraspMine}, including Direct and Supervised methods. 
Direct employs supervised learning with only one sample per object acquired from \textit{human-robot interaction}, while Supervised represents an upper-bound model trained with nearly 9k annotated samples. 
In offline experiments focusing on object grounding, PGA outperforms the Direct model by an average absolute increment of approximately 30\%, and even demonstrates comparable results to the Supervised method. 
Notably, we observe performance improvements in offline experiments as the number of images in \textit{Reminiscence} increases. 
Transitioning to the online setting, we validate PGA's real-world applicability by employing a physical robot to execute \textsc{GraspMine}. 
To qualitatively conclude our analysis, we present visual examples from different phases of PGA, providing insights into its robust performance.

% Contributions (3 itmes)
To summarize, our contributions are mainly three-fold:      

\begin{itemize}
\item We introduce a novel personalized task scenario in LCRG called \textsc{GraspMine}, aimed at grounding and grasping personal objects based on personal indicators. 
This scenario fills a gap in current LCRG systems, which primarily rely on generic expressions, thus enhancing the intuitive nature of human-robot interactions.
\item We propose Personalized Grasping Agent (PGA) as a strong baseline for \textsc{GraspMine}. PGA learns to ground personal objects by leveraging information acquired from a single \textit{human-robot interaction, robot-object interaction}, and \textit{Propagation through Reminiscence}.
\item We provide comprehensive experimental validation of PGA's performance against baselines, showcasing its effectiveness in grounding personal objects. 
Additionally, we demonstrate the PGA's practical applicability through deployment on a real-world physical robot.

\end{itemize}

%% file: 02_related_works.tex
\section{RELATED WORK}

\subsection{Language-Conditioned Robotic Grasping (LCRG)}
%\noindent \textbf{Language-Conditioned Robotic Grasping (LCRG)} 
Language-Conditioned Robotic Grasping (LCRG) focuses on the robot's capability to grasp objects based on human instructions given in natural language.
% At the heart of LCRG is the task of Visual Grounding (VG), where the primary challenge is to accurately identify objects as dictated by humans.
Numerous studies~\cite{srfnlifrm, invigorate, ingress, lopfribhsr, cpaptbgl, avgrefrm, ivgorefhri, reiofitsf} have employed fully supervised training approaches to teach robots how to locate objects based on natural language instructions, making use of public datasets~\cite{refcoco, refcocog, visual_genome}.
However, these supervised learning methods reach their limits when robots have to recognize or interact with personal objects unseen during the training.
A study~\cite{iprwowusli} curated its own dataset comprising unique objects. 
Still, it depended on supervised learning, which necessitates exhaustive annotations of language instructions.
GVCCI~\cite{gvcci} introduced a paradigm shift by unveiling an unsupervised approach to adapt to the user's personal objects. 
While GVCCI enabled robots to automatically learn the visual nuances of personal objects, it remained constrained. 
The learning was heavily reliant on generic object categories~\cite{basic_level_category} and attributes inferred by pretrained classifier~\cite{bottomup}.
% A major limitation of such methodologies is their grounding of visual objects in basic-level categories. 
This forces non-expert users to use \textit{robot-centric} instructions, a pain point discussed in Sec.~\ref{sec:intro}.
Some studies~\cite{rorwcnlq,prograsp} delved deeper, augmenting robots' knowledge with semantic or intended meanings of objects, typified by phrases like ``\textit{I am thirsty}''.
While these studies enable robots to grasp user intentions, they also rely on fully-supervised learning that can not be applied to \textsc{GraspMine}, necessitating extensive annotations.
% Personalized Grasping Agent (PGA), through an efficient semi-supervised learning approach that only requires a single annotation for each objects augmented by a robot's physical manipulation capabilities and environmental visual perceptions, can learn both visual properties of the user's personal objects and their corresponding natural language descriptors with just a single user-robot interaction.
Unlike previous works, PGA can efficiently learn the user's personal objects and their corresponding language descriptors with just a single user-robot interaction.

\subsection{Personalization}
% \noindent \textbf{Personalization} has become an important factor within various domains of Machine Learning, including image captioning~\cite{personalized_captioning1,personalized_captioning2,personalized_captioning3}, image retrieval~\cite{personalized_retrieval1,personalized_retrieval2}, semantic segmentation~\cite{personalized_segmentation}, and recommendation systems~\cite{personalized_reccomendation1}. 

%\noindent \textbf{Personalization} 
Personalization has become an important factor within various domains of Machine Learning \cite{personalized_captioning1,personalized_captioning2,personalized_captioning3,personalized_retrieval1,personalized_retrieval2,personalized_segmentation,personalized_reccomendation1,personalized_reccomendation2}.
Personalization has also been studied in the field of Robotics, such as dressing assistance~\cite{personalized_dressing1,personalized_dressing2,personalized_dressing3} and tidy-up task~\cite{personalized_tidyup}.
While these works of Personalization focus on the user's personal preferences, our work focuses on transferring the user's personal knowledge, standing as a pioneering work in the realm of personalized LCRG.
Two studies~\cite{fluffy, meta_personalizing} align closely with our research objectives, focusing on learning visual and linguistic representations of personal objects. 
The work \cite{meta_personalizing} utilizes personal videos paired with transcripts to learn personal objects and their corresponding names, without the need for explicit human labeling. 
Yet, their focus lies on temporal grounding in videos, overlooking the crucial aspect of spatial grounding. 
This limitation renders their approach unsuitable for LCRG, where pinpointing the precise object location is imperative. 
The work \cite{fluffy} broadens this focus to include spatial grounding.
However, they assume that fine-grained annotations (\textit{e.g.}, segmentation masks) and personal object names are readily available. 
In contrast, PGA only requires a single human-robot interaction without any further annotations. 
So, non-expert users can easily be involved in collecting the training data due to the intuitive and streamlined interface.

% Personalization section written by mjjung
% Personalization has become an important factor within various domains of Machine Learning \cite{personalized_captioning1,personalized_captioning2,personalized_captioning3,personalized_retrieval1,personalized_retrieval2,personalized_segmentation,personalized_reccomendation1,personalized_reccomendation2,personalized_reccomendation3}. 
% Several studies have addressed personalization capabilities in robotics, such as dressing assistance~\cite{personalized_dressing1,personalized_dressing2,personalized_dressing3} and tidy-up task \cite{personalized_tidyup}.
% While \cite{fluffy, meta_personalizing} focus on learning visual and linguistic representations of personal objects, they are not suitable for personalized LCRG since their lack of spatial grounding ability.
% To the best of our knowledge, none of existing LCRG methods \cite{?} have considered personalization capabilities, and our work stands as a pioneering work in this field.
% Moreover, unlike previous LCRG methods, our framework only requires a single human-robot interaction without any human annotations, demonstrating its efficiency.

%% file: 03_method.tex
\begin{figure*}[t]
\centering
\includegraphics[width=\linewidth]{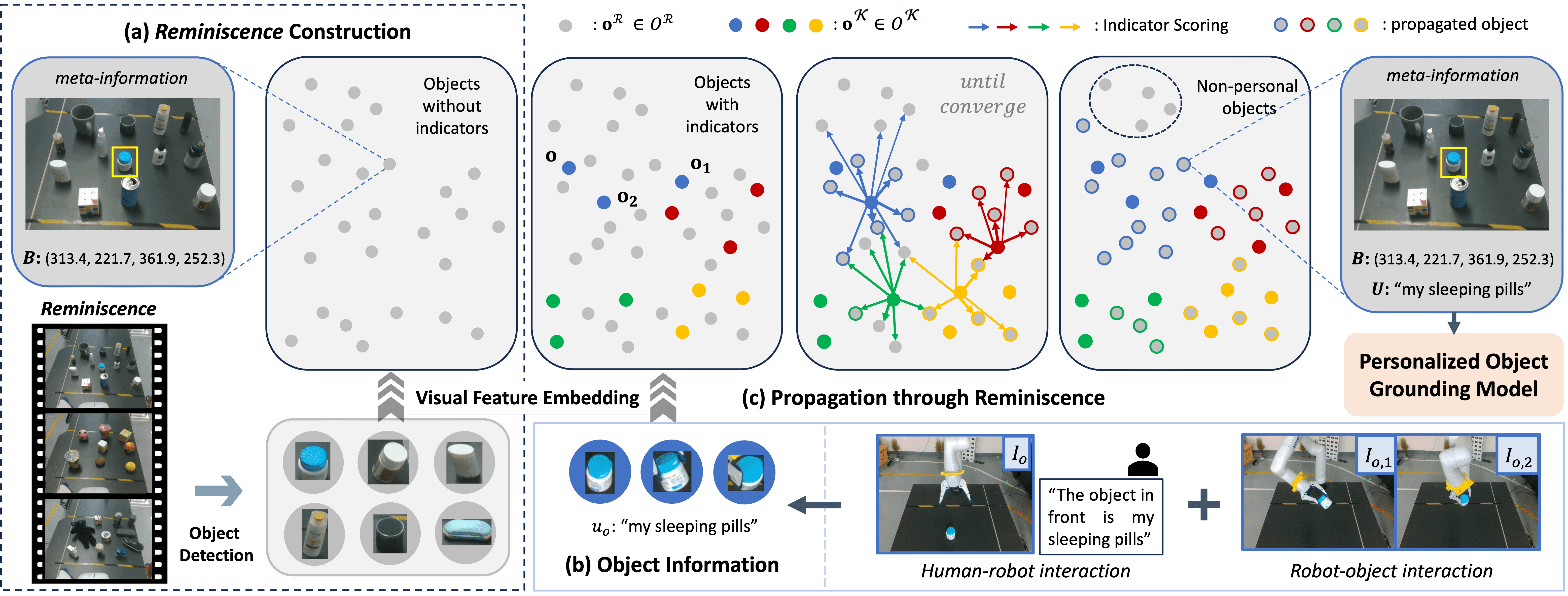}
\caption{\textbf{Overview of Personalized Grasping Agent (PGA).} 
(a) Initially, PGA gathers a collection of raw images, termed as \textit{Reminiscence}, from the user's environment. 
With the personal indicator provided by the user from (b), unlabeled objects in the \textit{Reminiscence} are pseudo-labeled via (c) Propagation through \textit{Reminiscence}. 
It's vital to note that certain objects, particularly those not introduced by the user (\textit{e.g.}, non-personal objects), remain unlabeled by our algorithm. 
Ultimately, PGA employs all the object nodes with labels (colored nodes) to train the Personalized Object Grounding Model.
}
\label{fig:main}
\vspace*{-0.3cm}
\end{figure*}

\section{METHOD}
\label{sec:method}

% The framework of Personalized Grasping Agent (PGA), is depicted in Fig.~\ref{fig:main}. 
% PGA is explained through a series of progressive steps, beginning with the PGA constructing the \textit{Reminiscence} (Sec.~\ref{ssec:remini_construction}), followed by acquiring information of the personal objects (Sec.~\ref{ssec:OIA}), learning processes of personal objects (Sec.~\ref{ssec:PR} and Sec.~\ref{ssec:grounding model}), and culminating in personalized object grasping (Sec.~\ref{ssec: personalized object grasping}). 

\subsection{Reminiscence Construction}
\label{ssec:remini_construction}
Personalized Grasping Agent (PGA) initiates by constructing the \textit{Reminiscence}, as illustrated in Fig.~\ref{fig:main}-(a).
PGA first collects a set of $N$ images from the user's environment, represented as $\mathcal{R}=\{I^\mathcal{R}_n\}_{n=1}^N$, which we term as \textit{Reminiscence}.
PGA detects all the objects $\{o^\mathcal{R}_m\}_{m=1}^M$ from $\mathcal{R}$, via an off-the-shelf object detector~\cite{bottomup}, and constructs a node for each detected objects, embedding them with the pretrained visual encoder $f(\cdot)$, DINO~\cite{dino}.
These object nodes, represented as a set of vector embeddings $O^\mathcal{R}=\{\mathbf{o}^\mathcal{R}_m\}_{m=1}^M=\{f(o^\mathcal{R}_m)\}_{m=1}^M$ where $M \gg N$, are tagged with the meta-information -- an image from $\mathcal{R}$ and the bounding box (bbox) coordinates of the detected object -- which will be leveraged in subsequent processes.
Note that nodes in $O^\mathcal{R}$ are \textit{without} the object's personal indicators.

\subsection{Object Information Acquisition}
\label{ssec:OIA}
To acquire the information of personal objects, PGA goes through two successive steps as shown in Fig.~\ref{fig:main}-(b): \textit{human-robot interaction} and \textit{robot-object interaction}.
In \textit{human-robot interaction}, the user introduces their personal object $o$ by displaying the object and verbally describing it.
The description consists of two indicators: a general indicator $u^G_o$, \textit{e.g.}, ``\textit{the object in front}'', and a personal indicator $u_o$, \textit{e.g.}, ``\textit{my sleeping pills}''.
Using its visual perception $I_o$ and $u^G_o$, PGA predicts the bbox coordinates $b_o$ of the object via GVCCI~\cite{gvcci}.
PGA then constructs an object node $\mathbf{o}=f(o)$ using pretrained visual encoder $f(\cdot)$~\cite{dino}.
This node is tagged with the meta-information including $I_o$, $b_o$, and $u_o$.

In \textit{robot-object interaction}, PGA subsequently examines the personal object $o$ from multiple views, aiming to capture its distinctive visual appearance.
By leveraging $b_o$, PGA grasps and rotates the object capturing $A$ images $\{I_{o,a}\}_{a=1}^A$ for each object, each image offering a distinct view of the object (see the bottom right of Fig.~\ref{fig:main}).
% As in the earlier step, PGA predicts an object bbox coordinates, $\{b_{o,a}\}_{a=1}^A$ of a set of objects $\{{o_a}\}_{a=1}^A$.
PGA extracts a set of objects $\{{o_a}\}_{a=1}^A$ and their bounding box coordinates $\{b_{o,a}\}_{a=1}^A$ from the captured images $\{I_{o,a}\}_{a=1}^A$. 
PGA constructs a set of object nodes $O^{view}=\{{\mathbf{o}_a}\}_{a=1}^A=\{{f(o_a)}\}_{a=1}^A$, where each node $\mathbf{o}_a$ is tagged with the meta-information $I_{o,a}$, $b_{o,a}$, and $u_o$.
Finally, each personal object is associated with $\mathbf{o}$ and $O^{view}$, \textit{i.e.}, $A+1$ nodes that are tagged with personal indicators. 
Across all personal objects, the aggregation of nodes with personal indicators forms $O^\mathcal{K}$, while collective personal indicators form $U$, as described in lines 1-6 of Algorithm~\ref{alg:labelpropagation}.
% Harnessing the robot's ability to manipulate objects, \textit{Multi-Angular Object Inspection} enables the robust recognition of the object under various viewing conditions in Sec.~\ref{ssec:PR} and~\ref{ssec:grounding model} as analyzed in Sec.~\ref{ssec:qualitative}.

% Written by mjjung
% \subsection{Multi-Angular Object Inspection}
% \label{ssec:MAOI}
% To capture an object's visual properties and distinct views, PGA grasps the object and rotates it into $N$ angular views to obtain object images $\{I_{o,n}\}_{n=1}^N$.
% Then, PGA crop each object image $I_{o,n}$ into $\check{{o}_n}$ and utilize a pretrained visual encoder $f(\cdot)$ to obtain object node $o_n$ as:
% \begin{align*}
% o_n = f(\check{o}_n) = f(Crop(I_{o,n})) \\
% \end{align*}
% Finally, we can the personal object nodes $O_{view} = \{o_n\}_{n=1}^{N}$ and each node have personal indicators $O^\mathcal{K}$.

\subsection{Propagation through Reminiscence}
\label{ssec:PR}

\textit{Propagation through Reminiscence} draws inspiration from the semi-supervised method of label propagation~\cite{label_propagation, label_propagation2}, as detailed in lines 7-20 of Algorithm~\ref{alg:labelpropagation} and Fig.~\ref{fig:main}-(c).
The goal is to pseudo-label (tag) the missing indicators in $O^\mathcal{R}$ by propagating personal indicators from $O^\mathcal{K}$.
For every node \(\mathbf{o}^\mathcal{R} \in O^\mathcal{R}\), PGA computes an affinity score, \(S(\mathbf{o}^\mathcal{R}, u)\), for each possible personal indicator $u \in U$. 
This score is the average cosine similarity $\phi$ between node \(\mathbf{o}^\mathcal{R}\) and every known object node \(\mathbf{o}^\mathcal{K}\) tagged with indicator \(u\);

\begin{equation}
\label{eq:cossim}
\begin{aligned}
\phi (\mathbf{o}^\mathcal{K},\mathbf{o}^\mathcal{R}) = \frac{\mathbf{o}^\mathcal{K} \cdot \mathbf{o}^\mathcal{R}}{\| \mathbf{o}^\mathcal{K} \|_2 \times \| \mathbf{o}^\mathcal{R} \|_2} = \frac{f(o^\mathcal{K}) \cdot f(o^\mathcal{R})}{\| f(o^\mathcal{K}) \|_2 \times \| f(o^\mathcal{R}) \|_2};
\end{aligned}
\end{equation}

\begin{equation}
% S(u) = \frac{1}{N(u)}\sum_{u_{o^\mathcal{K}} = u} \phi(o^\mathcal{R}, o^\mathcal{K});
% S(u) = \frac{1}{N(u)} \sum_{o^\mathcal{K} \in O^\mathcal{K}} I(u_{o^\mathcal{K}} = u) \cdot \phi(o^\mathcal{R}, o^\mathcal{K})
S(\mathbf{o}^\mathcal{R}, u) = \frac{1}{N(u)} \sum_{\mathbf{o}^\mathcal{K} \in O^\mathcal{K}} \mathbbm{1}(u_{\mathbf{o}^\mathcal{K}} = u) \cdot \phi(\mathbf{o}^\mathcal{R}, \mathbf{o}^\mathcal{K});
\end{equation}

% \begin{equation}
% S(u) = \frac{1}{N(u)}\sum_{\substack{o^\mathcal{K} \in O^\mathcal{K} \\ u_{o^\mathcal{K}} = u}} \phi(o^\mathcal{R}, o^\mathcal{K});
% \end{equation}

\noindent where $N(u)$ denotes the number of $\mathbf{o}^\mathcal{K}$s tagged with indicator $u$ and $\mathbbm{1}( \cdot )$ is the indicator function that takes the value of 1 if the condition is true, and 0 otherwise.
Note that $u_{\mathbf{o}^\mathcal{K}}, u \in U$.
Based on these scores, PGA assigns the indicator \(u\) with the highest affinity score to \(\mathbf{o}^\mathcal{R}\); $u_{o^\mathcal{R}}={\text{argmax}} \{ S(\mathbf{o}^\mathcal{R}, u) ~|~ u \in U \}$; if the highest score is above the certain threshold. 
After pseudo-labeling \(\mathbf{o}^\mathcal{R}\) with the most likely personal indicator,
$O^\mathcal{K}$ is updated to incorporate the propagated node \(\mathbf{o}^\mathcal{R}\), and label propagation repeats until the percentage of relabeled nodes is less than 10\%.
\textit{Propagation through Reminiscence} allows PGA to automatically gather pseudo-labeled examples, alleviating the scarcity of labeled data in \textsc{GraspMine}.
% fostering robust, data-rich learning of the user's personal object.

% \begin{equation}
% u_{o^\mathcal{R}} = \underset{u}{\mathrm{argmax}}\, S(u_{o^\mathcal{K}}) \quad \text{subject to} \quad \max_u S(u_{o^\mathcal{K}}) > \theta
% \end{equation}

% \begin{equation}
% \label{eq:cossim}
% \begin{aligned}
% $\phi (o',o'') = cossim(f(o'),f(o'')$)
% \end{aligned}
% \end{equation}

% Reminiscence graph $\mathcal{G}=(\mathcal{V},\mathcal{E})$

% $\mathcal{E}=[e_{i,j}]$ in which $e_{i,j}=cossim(f(v_i),f(v_j))$ for $v \gets \mathcal{V}$

\begin{algorithm}
\caption{Personalized Grasping Agent (PGA)}
\label{alg:labelpropagation}
\begin{algorithmic}[1]

% \Require A human user to interact with PGA
\Require The set of user's personalized objects $\mathcal{O}$
\Require Visual Encoder $f(\cdot)$~\cite{dino}
% \Require Initialize nodes $\mathcal{V}$ with unlabelled objects $O^R$
\Require Nodes $O^\mathcal{R} \gets \{\mathbf{o}^\mathcal{R}_m\}_{m=1}^M$ from $\mathcal{R}$ \Comment{(Sec.~\ref{ssec:remini_construction})}
\Require Set of personal indicators $U = \emptyset$
\Require Nodes with indicators $O^\mathcal{K} = \emptyset$

% $\check{o}^*$: Cropped bounding box image of $o^*$

\For{$o \gets \mathcal{O}$} \Comment{(Sec.~\ref{ssec:OIA})}
    % \State (Sec.~\ref{ssec:TPO})
    \State $\mathbf{o} \gets f(o)$ and $u_o$ from \textit{human-robot interaction}% \Comment{(Sec.~\ref{ssec:TPO})}
    \vspace{0.08cm}
    % \State $O^\mathcal{K} \gets O^\mathcal{K} \cup \{\mathbf{o}\}$ and $U \gets U \cup \{u_{o^{}}\}$ 
    % \For{$k \gets K$}
    \State Get $\{{o_a}\}_{a=1}^A$ from \textit{robot-object interaction}
    \vspace{0.08cm}
    \State $O^{view} \gets \{{f(o_a)}\}_{a=1}^A$ tagged with $u_o$  % \Comment{(Sec.~\ref{ssec:MAOI})}
    \vspace{0.08cm}
    \State $O^\mathcal{K} \gets O^\mathcal{K} \cup \{ \mathbf{o} \} \cup O^{view}$ and $U \gets U \cup \{u_{o^{}}\}$
    % \EndFor
\EndFor

% \State $\mathcal{V} \gets \mathcal{V} \cup O^\mathcal{K}$

\Repeat \Comment{(Sec.~\ref{ssec:PR})}
    \For{$\mathbf{o}^\mathcal{R} \gets O^\mathcal{R}$}
        \State $\forall u \in U$, set $S_u=0$, $N_u=0$
        
        \For{$\mathbf{o}^\mathcal{K} \gets O^\mathcal{K}$}
            \State $u \gets u_{\mathbf{o}^\mathcal{K}}$
            \State $S_u \gets S_u + \phi (\mathbf{o}^\mathcal{K},\mathbf{o}^\mathcal{R})$
            and $N_u \gets N_u + 1$
            
        \EndFor
        % \For{$u \gets U$}
        \State $\forall u \in U$, $S(\mathbf{o}^\mathcal{R}, u) \triangleq S_u/N_u$
        % \EndFor
        % \State indicator-wise vote averaging
        \If{max $\{ S(\mathbf{o}^\mathcal{R}, u) ~|~ u \in U \}$ $>$ threshold}
            % \State set the label of $o^\mathcal{R}$ with max voted $l \in U$
            \State $u_{o^\mathcal{R}} \gets \text{argmax} \{ S(\mathbf{o}^\mathcal{R}, u) ~|~ u \in U \}$
            \vspace{0.08cm}
            \State $O^\mathcal{K} \gets O^\mathcal{K} \cup \{\mathbf{o}^\mathcal{R}\}$
        \EndIf
    \EndFor
\Until{Ratio of changed indicator of node $<$ 10\%}

\end{algorithmic}
\end{algorithm}

% \subsection{Identifier Diversification}

% The technique of \textit{Identifier Diversification} significantly enhances the repertoire of personal object indicators. 
% We harness the capabilities of the Large Language Model (LLM), GPT-3~\cite{gpt3,chagpt}, to diversify these indicators. 
% We formulate the input prompt in the following manner: ``Please paraphrase the phrase \{$u_o$\} using \{$n$\} different expressions that convey the same meaning''.
% The diversified indicators generated using this approach are then employed to train the grounding model (Sec.~\ref{ssec:grounding model}), thereby promoting a more holistic and flexible understanding of personal object indicators.

\subsection{Personalized Object Grounding Model}
\label{ssec:grounding model}

The \textit{Personalized Object Grounding Model} is a Transformer~\cite{transformer} based Vision-Language model that takes an image and a natural language indicator to infer the bounding box coordinates of the personal object. 
% We adopt the architecture of OFA~\cite{ofa}, a Transformer~\cite{transformer} based Vision-Language model. 
The model's optimization occurs by minimizing 

\begin{equation}
\label{eq:loss}
\begin{aligned}
\mathcal{L}=-\sum_{\mathbf{o}^\mathcal{K} \in O^\mathcal{K}}logP_{\theta}(b_{\mathbf{o}^\mathcal{K}}|I_{\mathbf{o}^\mathcal{K}},u_{\mathbf{o}^\mathcal{K}}),
\end{aligned}
\end{equation}

\noindent the negative log-likelihood of the bounding box for every object node $\mathbf{o}^\mathcal{K} \in O^\mathcal{K}$.
Note that $O^\mathcal{K}$ is not from Sec.~\ref{ssec:OIA} but after \textit{Propagation through Reminiscence} (Sec.~\ref{ssec:PR}).

\subsection{Personalized Object Grasping}
\label{ssec: personalized object grasping}

The process of \textit{Personalized Object Grasping} is initiated when a user instructs the robot to grasp a personal object as seen in Fig.~\ref{fig:first}. 
To execute this, PGA first infers the 2D coordinates of the queried object using the \textit{Personalized Object Grounding Model}. 
Upon obtaining these 2D bounding box coordinates, PGA calculates the 3D segmented object coordinates by leveraging point cloud data and the RANSAC~\cite{ransac} algorithm. 
Specifically, it translates the 2D bounding box into a 3D spatial configuration using the point cloud, followed by segmenting the points of the object within the 3D bounding box via RANSAC. 
Lastly, PGA grasps the object by computing a trajectory of the robot arm~\cite{moveit}.

%% file: 04_dataset.tex
\begin{table*}[t]
\caption{
\begin{flushleft}
\textnormal{\textbf{Results of offline experiments.}
Results show the personal object grounding score in \textsc{GraspMine} dataset. 
Bold scores represent the highest performance, while underlined scores the second-best results.
The `interaction' column represents the average number of interactions per object that users had with the robot to teach personal objects. 
`annotated' indicates the total human annotations used in each method, while `utilized' specifies the number of triplets - image, object bounding box, and personal indicator - employed for training the models. 
It's noteworthy that subtracting the `annotated' count from the `utilized' count reveals the amount of triplet automatically generated by each method without any human labor.
}
\end{flushleft}
}
  \centering
  \resizebox{\textwidth}{!}{
  \begin{tabular}{lccccccccccc}
    \hline
    \toprule
    \multirow{2}{*}{} & \multirow{2}{*}{} & \multirow{2}{*}{} & \multirow{2}{*}{} & \multicolumn{2}{c}{Heterogeneous} & \multicolumn{2}{c}{Homogeneous} & \multicolumn{2}{c}{Paraphrased} & Cluttered & Generic \\ 
    \cmidrule(lr){5-6}\cmidrule(lr){7-8}\cmidrule(lr){9-10}\cmidrule(lr){11-11}\cmidrule(lr){12-12}   
    Method & interaction & annotated & utilized & $\text{\footnotesize IoU}_{>0.5}$ & $\text{\footnotesize IoU}_{>0.8}$ & $\text{\footnotesize IoU}_{>0.5}$ & $\text{\footnotesize IoU}_{>0.8}$ & $\text{\footnotesize IoU}_{>0.5}$ & $\text{\footnotesize IoU}_{>0.8}$ & $\text{\footnotesize IoU}_{>0.8}$ & $\text{\footnotesize IoU}_{>0.5}$ \\
    \midrule
    OFA~\cite{ofa} & - & - & - & 49.2 & 47.5 & 23.7 & 20.3 & 35.8 & 34.1 & 34.6 & 65.7 \\
    GVCCI~\cite{gvcci} & - & - & - & 59.3 & 55.1 & 30.5 & 23.8 & 42.3 & 38.9 & 44.3 & \underline{79.1}\\
    \midrule
    Direct & 1.1 & 96 & 96 & 60.2 & 55.1 & 37.3 & 27.1 & 48.3 & 42.5 & 46.4 & - \\
    PassivePGA & 1.1 & 96 & 4,828 & 89.0 & 76.3 & 68.6 & 53.4 & 73.6 & 62.7 & 62.4 & - \\
    PGA (ours) & 1.1 & 96 & 6,492 & \underline{91.5} & \underline{81.4} & \underline{70.3} & \underline{61.9} & \underline{74.7} & \underline{68.2} & \underline{64.5} & \underline{79.1}\\
    \midrule
    Supervised & 91.3 & 8,763 & 8,763 & \textbf{97.5} & \textbf{90.7} & \textbf{92.4} & \textbf{83.1} & \textbf{84.8} & \textbf{79.0} & \textbf{71.4} & - \\
    \bottomrule
    \hline
  \end{tabular}}
  \label{tab:offline}
  \vspace*{-0.3cm} 
\end{table*}

\section{GraspMine}
\label{sec:graspmine}

Our proposed task scenario, \textsc{GraspMine}, consists of a curated dataset comprising a Training set, \textit{Reminiscence}, and Test set, featuring 96 personal objects along with 100+ everyday objects, totaling around 200 individual objects.

\noindent \textbf{Training set} consists of 96 pairs of images $I_o$ and their respective personal indicators $u_o$, alongside general indicators $u_o^G$, collected through \textit{human-robot interaction} described in Sec.~\ref{ssec:OIA} and illustrated in Fig.~\ref{fig:main}-(b).

\noindent \textbf{\textit{Reminiscence}}, $\mathcal{R}=\{I_n^\mathcal{R}\}_{n=1}^N$, involves $N=400$ raw images, each containing multiple objects.
These unlabeled images can be optionally provided to aid the learning process.
However, we annotated objects in the \textit{Reminiscence} for analyzing the propagation ability of models and for training the Supervised method in Sec.~\ref{ssec:offline}.
Notably, these annotations were not utilized in any aspect of training PGA.

\noindent \textbf{Test set.} 
% The test set aims to validate the agent's ability to infer the bounding box coordinates of personal objects given a personal indicator learned from the training process.
Examples of a Test set is presented in Fig.\ref{fig:qualitative}-A$\sim$E. 
Given an image and a personal indicator, the agent is tasked with inferring the correct location (depicted by the black box).
For the deeper analysis, Test set is categorized into five distinct splits as follows.
\begin{itemize}
\item \textbf{Heterogeneous} split incorporates scenes with randomly selected objects as exemplified in Fig.\ref{fig:qualitative}-A. 
It consists of 60 images with 120 personal indicators and bboxes.
\item \textbf{Homogeneous} split incorporates scenes with similar-looking objects with the same category, making discrimination more challenging as exemplified in Fig.\ref{fig:qualitative}-B,C.
It consists of 60 images with 120 indicators and bboxes.
\item \textbf{Cluttered} split contains 106 images containing highly cluttered objects as exemplified in Fig.\ref{fig:qualitative}-D,E, and a single personal indicator and bbox per image. Images of this split originate from the IM-Dial dataset~\cite{prograsp}.
\item \textbf{Paraphrased} split comprises all Heterogeneous, Homogeneous, and Cluttered split with each personal indicator paraphrased by the annotators. For instance, the personal indicator ``\textit{my sleeping pills}'' is queried as ``\textit{the medication I take for sleep}'' as depicted in Fig.~\ref{fig:first}.
\item \textbf{Generic} split is sourced from the \textit{VGPI} dataset presented in GVCCI~\cite{gvcci}.
This dataset features images comprising generic objects, each annotated with indicators based on their basic-level object categories. 
Within the \textit{VGPI} dataset, we specifically utilized the Test-E split, which aligns with our environmental settings.
\end{itemize}

% Scenes in all the splits also contain objects that were unseen during training.

%% file: 04_experiment.tex
\section{EXPERIMENTS}
% We assessed the Personalized Grasping Agent (PGA) on our proposed dataset, \textsc{GraspMine}, benchmarking it against various baselines.
% The offline experiment measured PGA's efficacy in Personalized Object Grounding, how well PGA identifies an object given its natural language indicators. 
% Meanwhile, the online experiment probed its real-world performance in Personalized Language-Conditioned Robotic Grasping (LCRG) using a robot arm.

\subsection{Compared Methods}
% First, we compared PGA against previous methods, considering its specific training setup.
\noindent \textbf{OFA}~\cite{ofa} is a state-of-the-art visual grounding model pre-trained using a public dataset~\cite{refcoco}. 
This model has never been exposed to the user's personal objects, positioning it as the most basic compared method. 
It primarily relies on generic cues (\textit{e.g.}, “\textit{tennis ball}”) present in personal indicators (\textit{e.g.}, “\textit{my Djokovic tennis ball}”) to make predictions.

\noindent \textbf{GVCCI}~\cite{gvcci} is a robotic lifelong learning framework designed to autonomously learn generic expressions of objects. By leveraging images from both the training set and the \textit{Reminiscence}, it automatically generates these generic expressions and utilizes the generated data for training.
% It shares the same model architecture with OFA.
%This model harnesses the power of the \textit{Reminiscence} dataset, comprising personal objects. However, this method does not account for personal indicators, training the model only with generic expressions.

% To verify the importance of each model component, we conduct experiments with several variants of PGA.
\noindent \textbf{Direct} model is trained directly from the \textit{human-robot interaction} (Sec.~\ref{ssec:OIA}). 
This model does not employ \textit{robot-object interaction} (Sec.~\ref{ssec:OIA}) and the \textit{Propagation through Reminiscence} (Sec.~\ref{ssec:PR}), resulting in training the model with a single sample per personal object. 
This model serves as a representation of traditional LCRG systems that typically learn in a supervised manner when confronted with the task scenario of \textsc{GraspMine}. 
Consequently, the Direct model can only utilize 96 annotated data obtained from an average of 1.1 \textit{human-robot interaction} per object (101 interactions for 96 objects due to five visual grounding failures).

\noindent \textbf{PassivePGA} model is an ablative model of PGA, excluding \textit{robot-object interaction} in Sec.~\ref{ssec:OIA}. 
In other words, \textit{Propagation through Reminiscence} (Sec.~\ref{ssec:PR}) is conducted only with the nodes from \textit{human-robot interaction} in Sec.~\ref{ssec:OIA}.
Consequently, this model utilizes 96 annotated data points along with 4.7k pseudo-labeled data obtained from the \textit{Propagation through Reminiscence}, resulting in a total of 4.8k utilized data points.

\noindent \textbf{Supervised} model is trained with a fully-supervised approach that utilizes ground-truth annotations (\textit{i.e.}, object location coordinates and personal indicators) in \textit{Reminiscence}. 
Since this model has access to all 8.7k ground truth annotations in the \textit{Reminiscence}, whereas \textsc{GraspMine} assumes access of only one annotation per object (96 annotations in total), the Supervised model serves as the upper bound performance benchmark for \textsc{GraspMine} in the following experiments. Essentially, training this model requires approximately 100 times more manual effort compared to PGA.

\subsection{Offline Experiment}
\label{ssec:offline}

\noindent \textbf{Evaluation Protocol.} 
In the offline experiment, we study the PGA's proficiency in grounding the target object given a personal indicator.
By following standard practice~\cite{gvcci,ofa}, we assess the grounding accuracy using the Intersection over Union (IoU) score, a key visual grounding metric, which calculates the overlap between the predicted and ground truth bounding boxes. 
While we present the percentage of predictions surpassing an IoU of 0.5, we also emphasize IoU above 0.8 to ensure a stricter alignment between the predicted and actual regions, acknowledging the precision needed for successful object grasping.
For the Cluttered split, we only reported scores with an IoU exceeding 0.8 since boxes with an IoU above 0.5 frequently included multiple objects, rendering the score less meaningful.

\noindent \textbf{Comparison with baselines.}
The offline results, presented in Tab.~\ref{tab:offline}, demonstrate that the Direct model, which relies solely on object information acquired from \textit{human-robot interaction} without leveraging \textit{Reminiscence}, exhibited only marginal improvement compared to methods reliant on generic object knowledge, \textit{i.e.}, OFA and GVCCI. 
This indicates that naively training existing LCRG models on a small amount of labeled data does not make significant performance gains in \textsc{GraspMine}.
However, PGA, utilizing a large amount of unlabeled data, demonstrated a significant improvement compared to the Direct model, achieving approximately 30\% enhancement.
Remarkably, even when compared to the Supervised method, which requires about 100 times more annotations than PGA, PGA exhibited comparable results. 
We also assess PGA in grounding objects when queried with generic instructions, using the Generic split, to examine how well PGA retains knowledge about the generic instructions after training the personal indicators. 
Even after acquiring personal knowledge, PGA's ability to ground objects through generic instructions remains impeccably intact.
In short, by utilizing an \textit{Reminiscence} and the robot's manipulative power, PGA proficiently grounds personal objects with learning from a single human-robot interaction, outperforming the baseline methods and showing comparable performance to the Supervised approach, all while retaining its knowledge of generic expressions.

\begin{figure*}[t]
\centering
\includegraphics[width=\linewidth]{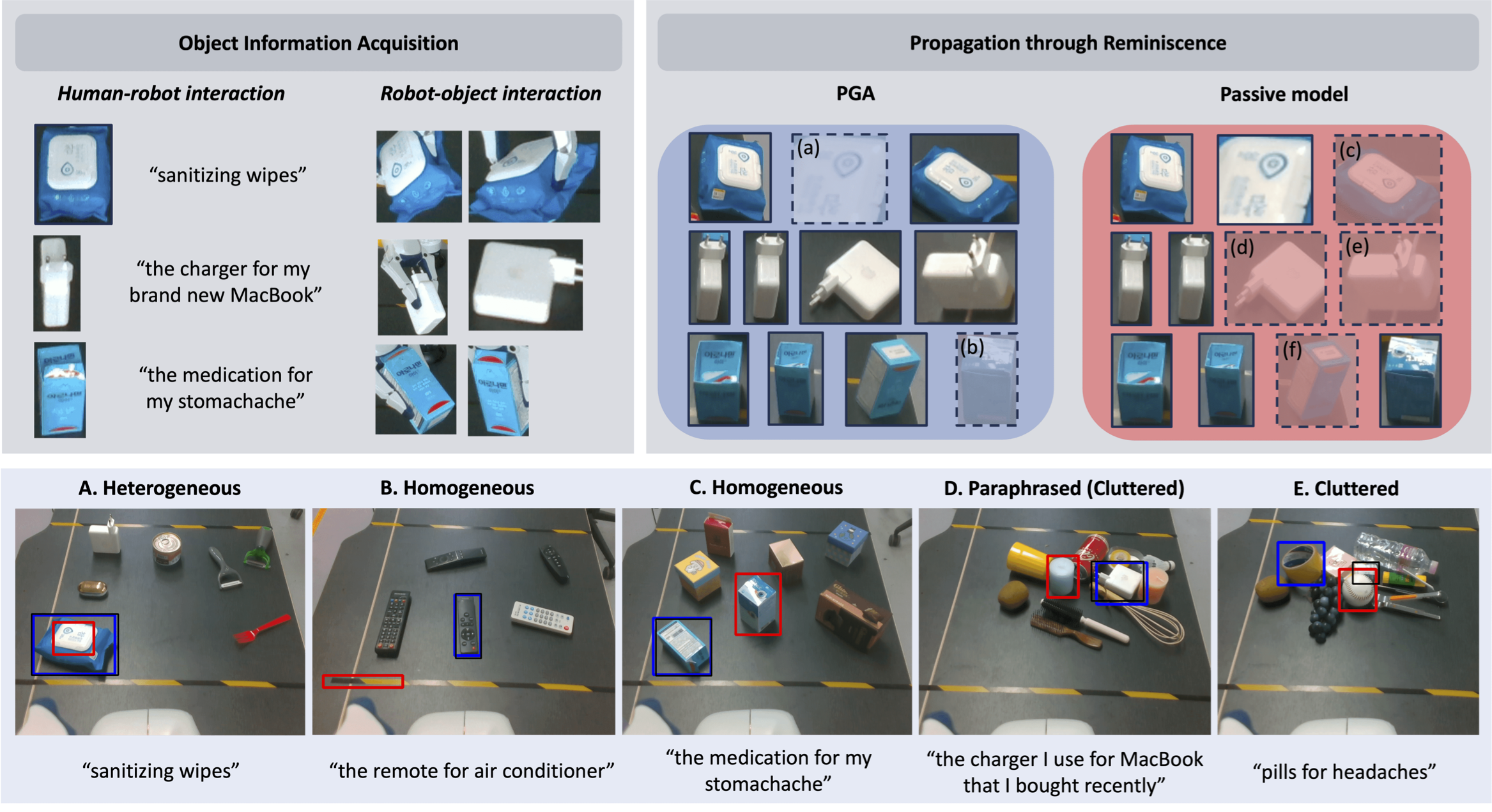}
\caption{
\textbf{Qualitative Analysis.} 
The top row showcases examples of objects from each phase. In the bottom row, personalized object grounding results by PGA are depicted in blue boxes and those by PassivePGA model in red, set alongside the ground truth in black boxes. Within the \textit{Propagation through Reminiscence} on the top row, solid lined boxes are the pseudo-labeled objects from the \textit{Reminiscence} and dotted boxes denote objects that were NOT pseudo-labeled according to indicators in the respective models.
}
\label{fig:qualitative}
\vspace*{-0.2cm}
\end{figure*}

\noindent \textbf{Impact of \textit{robot-object interaction}.}
Tab.~\ref{tab:offline} shows a notable enhancement of up to 8.5\% in PGA over the PassivePGA model.
This underscores the impact of \textit{robot-object interaction} in improving performance by capturing objects from different views. 
Notably, when comparing the $IoU_{>0.5}$ and $IoU_{>0.8}$ across various splits in Tab.~\ref{tab:offline}, we found that the improvement in $IoU_{>0.8}$ (+5-8\%) exceeded that of $IoU_{>0.5}$ (+1-2\%). 
This suggests that leveraging \textit{robot-object interaction} leads to more precise localization of objects, particularly beneficial for robotic grasping. 
% Further comprehensive analysis of these findings will be provided in the ``Analysis of \textit{Propagation through Reminiscence}'' section.
To further understand these findings, we compare the propagation accuracy of PassivePGA and PGA.
The accuracy, measured as the percentage of correctly pseudo-labeled objects among total nodes, revealed PGA achieving 79.4\%, while PassivePGA lagged behind at 64.5\%.
We further investigate incorrectly labeled objects and observe that most of the propagation error comes from the noisiness of the off-the-shelf object detector~\cite{bottomup}.
Noisy objects were categorized into two groups: 1) ambiguous boxes, \textit{i.e.}, boxes that only include a part of an object (\textit{e.g.}, Fig.~\ref{fig:qualitative}-(a)) or boxes that surround multiple objects, and 2) invalid boxes, \textit{i.e.}, non-objects such as line stickers on desks or a robot arm (\textit{e.g.}, Fig.~\ref{fig:qualitative}-B).
We found 393 ambiguous boxes and 870 invalid boxes in the \textit{Reminiscence}. 
% We compare PGA with the PassivePGA model that does not have \textit{robot-object interaction}. 
Regarding propagation to ambiguous boxes, PGA showed a reduced rate of 41.2\% compared with that of 50.4\% from the PassivePGA model. 
Moreover, PassivePGA model had a high tendency to propagate towards invalid boxes, at 83.7\%, while PGA showed a nearly negligible rate at 0.07\%.
In summary, the impact of \textit{robot-object interaction} on the inaccurate and noisy \textit{Propagation through Reminiscence} appears to have influenced both the overall grounding accuracy and the precision of localization. 
% \noindent \textbf{Impact of \textit{robot-object interaction}.}

\begin{figure}[t]
\centering
\includegraphics[width=0.94\linewidth]{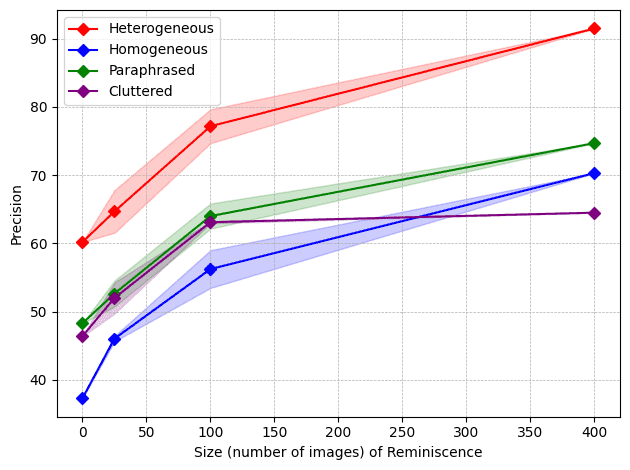}
\caption{
\textbf{Impact of \textit{Reminiscence} size.} 
PGA's offline scores based on the number of raw images utilized from \textit{Reminiscence}, from `0' to `400'.
}
\label{fig:remini_result}
\vspace*{-0.3cm}
\end{figure}

\noindent \textbf{Impact of \textit{Reminiscence} size.}
As shown in Fig.~\ref{fig:remini_result}, our exploration delves into the relationship between PGA's performance and the size of the \textit{Reminiscence} — the number of raw images utilized. 
A size of `0' is equivalent to the Direct model. 
We experimented on a log scale, examining sizes of 25, 100, and 400, with each configuration undergoing three separate trials. 
Our findings suggest a clear pattern: as PGA is exposed to an increasing number of scenes from the user's environment, its performance is incrementally enhanced.
This indicates that, as the robot collects more raw images from the environment, its performance is expected to be improved without necessitating further human supervision.

% \noindent \textbf{Analysis of Propagation through Noisy \textit{Reminiscence}.}
% Detected objects in the \textit{Reminiscence} are noisy as they are predicted using an off-the-shelf object detector~\cite{fasterRCNN}.
% Therefore, in this analysis, we investigate how many noisy objects are pseudo-labelled with Propagation through \textit{Reminiscence}.
% We divided noisy objects into two categories: 1) ambiguous boxes, \textit{i.e.}, boxes that only include a part of an object (\textit{e.g.}, Fig.~\ref{fig:qualitative}-(a)) or boxes that surround multiple objects, and 2) invalid boxes, \textit{i.e.}, non-objects such as line stickers on desks or a robot arm (\textit{e.g.}, Fig.~\ref{fig:qualitative}-B).
% We found 393 ambiguous boxes and 870 invalid boxes in the \textit{Reminiscence}. 
% We compare PGA with the PassivePGA model that does not have \textit{robot-object interaction}. 
% Regarding propagation to ambiguous boxes, PGA showed a reduced rate of 41.2\% compared with that of 50.4\% from the PassivePGA model. 
% Moreover, PassivePGA model had a high tendency to propagate towards invalid boxes, in at 83.7\%, while PGA showed nearly negligible rate at 0.07\%.

\subsection{Online Experiment}

\noindent \textbf{Robotic Platform.} 
Our robotic platform is equipped with the 6-DoF Kinova Gen3 Lite\footnote{https://www.kinovarobotics.com/product/gen3-lite-robots}, complemented by the Intel Realsense Depth Camera D435. While all learning and inferences of PGA are managed on the remote server, manipulation planning is computed on the local platform.

\noindent \textbf{Evaluation Protocol.}
From the Heterogeneous and Homogeneous splits, we randomly selected 30 images each and closely reproduced the depicted scenes. 
We assessed LCRG based on two human-evaluated metrics: object grounding success rate and grasping accuracy, \textit{i.e.}, LCRG.

\noindent \textbf{Results.}
Results in Tab.~\ref{tab:online} reveal that PGA significantly surpassed the Direct model in grasping, along with enhanced grounding accuracy. 
The Direct model, which solely relies on object information acquired from \textit{human-robot interaction} without leveraging raw images from \textit{Reminiscence}, lagged behind PGA with a performance gap of 23.4\% in the Heterogeneous split and 33.3\% in the Homogeneous split. 
This result underscores the effectiveness of PGA, which achieves superior performance without requiring additional human labor for annotations compared to the Direct model.
% Notably, PGA's performance was comparable to the Supervised method, which necessitates extensive labeling of raw \textit{Reminiscence} images, highlighting our method's efficacy with only requiring single human-robot interaction to transfer personal knowledge.

% \begin{table}[t]
% \caption{
% \begin{flushleft}
% \textnormal{\textbf{Results of Online Experiment (LCRG)} in \textsc{GraspMine}.
% % Bold scores represent the highest performance, while underlined scores the second-best results.
% }
% \end{flushleft}
% }
%   \centering
%   \resizebox{0.48\textwidth}{!}{
%   \begin{tabular}{lcccc}
%     \hline
%     \toprule
%     \multirow{2}{*}{} & \multicolumn{2}{c}{Heterogeneous} & \multicolumn{2}{c}{Homogeneous} \\ 
%     \cmidrule(lr){2-3}\cmidrule(lr){4-5}  
%     Method & grounding & grasping & grounding & grasping \\
%     \midrule
%     Direct & 63.3 & 53.3 & 26.7 & 20.0 \\
%     PGA (ours) & \underline{90.0} & \underline{76.7} & \underline{66.7} & \underline{53.3} \\
%     \midrule
%     Supervised & \textbf{100.0} & \textbf{83.3} & \textbf{83.3} & \textbf{70.0} \\
%     \bottomrule
%     \hline
%   \end{tabular}}
%   \label{tab:online}
% \vspace*{-0.2cm}
% \end{table}

\begin{table}[t]
\caption{
\begin{flushleft}
\textnormal{\textbf{Results of Real-world Online Experiment (LCRG)} in \textsc{GraspMine}.
% Bold scores represent the highest performance, while underlined scores the second-best results.
}
\end{flushleft}
}
  \centering
  \resizebox{0.48\textwidth}{!}{
  \begin{tabular}{lcccc}
    \hline
    \toprule
    \multirow{2}{*}{} & \multicolumn{2}{c}{Heterogeneous} & \multicolumn{2}{c}{Homogeneous} \\ 
    \cmidrule(lr){2-3}\cmidrule(lr){4-5}  
    Method & grounding & grasping & grounding & grasping \\
    \midrule
    Direct & 63.3 & 53.3 & 26.7 & 20.0 \\
    PGA (ours) & \textbf{90.0} & \textbf{76.7} & \textbf{66.7} & \textbf{53.3} \\
    \bottomrule
    \hline
  \end{tabular}}
  \label{tab:online}
%\vspace*{-0.2cm}
\end{table}

\subsection{Qualitative Analysis}
\label{ssec:qualitative}
Through qualitative analysis, shown in Fig.~\ref{fig:qualitative}, we visualize examples from the various phases including \textit{Object Information Acquisition}, \textit{Propagation through Reminiscence}, and \textit{Personalized Object Grounding}.
For a comprehensive understanding, we compare the performance of PGA against the PassivePGA model.
% The PassivePGA model does not leverage the data acquired from \textit{robot-object interaction}.

One striking observation is PGA's adeptness at inspecting objects from diverse angles as seen in `\textit{robot-object interaction}'. 
This allows PGA to recognize variations in object orientation, \textit{e.g.}, views from the rear or the base. 
In instances such as those represented in (c), (d), (e), and (f), the PassivePGA model often overlooks objects in the \textit{Reminiscence} that differ from their appearances in the \textit{human-robot interaction} phase. 
We conjecture that the omitted samples, \textit{e.g.}, (d), (e), and (f), in the training phase affect some wrong prediction results as in C and D.
For example, consider the case of ``the charger for my brand new MacBook''. 
With the aid of \textit{robot-object interaction}, PGA successfully propagates to samples viewed from different angles. 
However, PassivePGA fails to capture such samples as shown in (d) and (e). 
In sample D, it is evident that PassivePGA makes an incorrect inference, while PGA's inference is accurate.
%This oversight manifests in the form of incorrect inferences, particularly evident in panels C and D.
Moreover, as detailed in Sec.~\ref{ssec:offline}, the PassivePGA model occasionally tends to propagate towards ambiguous boxes in the \textit{Reminiscence}, as depicted in (a). 
Such behavior can lead to inaccurate grounding, as observed in the red box in sample A. 
The tendency of propagating towards incorrect boxes might have affected in some results shown in panel B.

% Further, the PassivePGA model occasionally displays a tendency to propagate towards ambiguous objects in Reminiscence. 
% This phenomenon is illustrated in (a), which may result in inaccurate grounding as seen in A.
% Recap, that since the object bounding boxes in the \textit{Reminiscence} is the ones detected from the off-the-shelf object detector, the boxes might be noisy.
% Annotating the object bounding boxes in the \textit{Reminiscence}, we found out that there were 393 ambiguous boxes like (a) and 870 wrong boxes such as line stickers on desks as seen in the panel B.
% Measuring the rate of wrong propagation towards ambiguous objects within \textit{Reminiscence}, PGA exhibited a comparatively lesser rate of 41.2\%, as opposed to the PassivePGA model's 50.4\%.
% Another aspect found is the propagation to wrong objects in the environment such as line stickers on desks. 
% While the PassivePGA model exhibited a high tendency to propagate towards these objects, in at 83.7\%, PGA's tendency was nearly negligible at 0.07\%. This resulted in some cases shown in B.

Despite PGA's robust grounding abilities as seen through A-D, challenges persist in some complex scenarios, such as in a scenario like E, where objects are situated in highly cluttered environments---the difficulty reflected in the lower scores of Cluttered split in Tab.~\ref{tab:offline}. 
Nevertheless, our qualitative analysis reveals PGA's successful \textit{Propagation through} \textit{Reminiscence}, enhanced by the \textit{robot-object interaction}, yielding promising results in Personalized Object Grounding.

%% file: 05_discussion.tex
\section{DISCUSSION}
\label{sec:discussion}

The Personalized Grasping Agent (PGA) has shown promising performance in the \textsc{GraspMine} dataset. However, it is essential to discuss several assumptions and limitations in our experimental setup to guide future studies for better improvement in dealing with \textsc{GraspMine}.

Due to the use of an arm robot without mobility necessitates two primary assumptions. 
Firstly, we confine the execution area to a 80*60 \textit{cm} table where all objects are at least partially visible within this space.
This assumption limits manipulation to flat and open surfaces, excluding scenarios such as retrieving objects from handbags or shelves. 
Overcoming this limitation would require developing deeper and dynamically adaptive reasoning in motion planning, a direction we leave for future exploration.
Secondly, we assume a full access to raw images of \textit{Reminiscence}. 
While obtaining these raw images from a mobile robot through random navigation in the user's environment is feasible, we assumed that these images were pre-acquired. 
Extending this work with a mobile robot capable of autonomously acquiring raw images from the user’s environment would be a crucial topic for further studies.

% Limitations stemming from the experimental results include:

% Discrepancies observed in the offline experiment (See Tab.~\ref{tab:offline}) between PGA and the Supervised method was particularly evident (22\% performance gap) in Homogeneous split, where similar-looking objects are present. 
% This disparity was due to imperfect accuracy in the "\textit{Propagation through Reminiscence}," \textit{i.e.}, mispredictions when identifying identical objects in the \textit{Reminiscence}. Thus, we believe that developing a sophisticated label propagation algorithm to reduce the number of noisy labels will be an interesting direction for future work.

During the online experiment with the physical robot (See Tab.~\ref{tab:online}), a performance drop from grounding accuracy to grasping accuracy (around 13\%) was observed. This decline can be attributed to limitations in motion planning, particularly in the manipulation strategy employed, which first aligns the object and the end effector's horizontal coordinates before making vertical adjustments. However, this approach may prove ineffective when attempting to grasp objects like wine bottles. Enhancing motion planning to tailor grasping strategies for diverse objects is crucial for future studies.

% Addressing these assumptions and limitations will be vital for enhancing the effectiveness and applicability of PGA in real-world scenarios, ultimately advancing the field of personalized Language-Conditioned Robotic Grasping.

%% file: 06_conclusion.tex
\section{CONCLUSION}
This paper introduces a challenging task scenario \textsc{GraspMine}, along with a novel robotic framework PGA to equip robots with the capacity to learn a user's personalized knowledge through a single human-robot interaction. 
By expanding the mutual knowledge between a robot and its user beyond the generic, our approach promotes a shift from \textit{robot-centric} to \textit{user-centric} interactions. 
We anticipate that our method will foster more natural, intuitive interaction and significantly enhance the non-expert user’s experience.